\documentclass[10pt]{article}
\usepackage[letterpaper,margin=1in]{geometry}
\usepackage[T1]{fontenc}
\usepackage[utf8]{inputenc}
\usepackage{newunicodechar}
\usepackage{times}
\usepackage{amsmath,amssymb}
\usepackage{graphicx}
\usepackage{booktabs,longtable,array,calc}
\usepackage{etoolbox}
\usepackage{caption}
\usepackage[hidelinks]{hyperref}
\usepackage{microtype}
\newunicodechar{§}{\S{}}
\newunicodechar{±}{\ensuremath{\pm}}
\newunicodechar{·}{\ensuremath{\cdot}}
\newunicodechar{×}{\ensuremath{\times}}
\newunicodechar{α}{\ensuremath{\alpha}}
\newunicodechar{β}{\ensuremath{\beta}}
\newunicodechar{ε}{\ensuremath{\varepsilon}}
\newunicodechar{η}{\ensuremath{\eta}}
\newunicodechar{σ}{\ensuremath{\sigma}}
\newunicodechar{∂}{\ensuremath{\partial}}
\newunicodechar{−}{\ensuremath{-}}
\newunicodechar{≥}{\ensuremath{\geq}}
\newunicodechar{⊙}{\ensuremath{\odot}}
\providecommand{\tightlist}{\setlength{\itemsep}{0pt}\setlength{\parskip}{0pt}}

\AtBeginEnvironment{longtable}{\small}
\AtBeginEnvironment{verbatim}{\footnotesize}
\title{More Bits Won't Fix It: Diagnosing and Treating an Ill-Conditioned Backward Pass (A Case Study on Gemma-4)}
\author{Jeong Hyun Lee, MD\\[0.3em]
\small Department of Radiology and Center for Imaging Science,\\
\small Samsung Medical Center, Sungkyunkwan University School of Medicine,\\
\small 81 Irwon-ro, Gangnam-gu, Seoul 06351, Republic of Korea}
\date{Preprint, November 2026}

\begin{document}
\maketitle
\begin{abstract}
During LoRA fine-tuning of Gemma-4 (12B and 31B) on long prompts, whether they contain CT images, natural photos,
or about 2,000 tokens of text, bf16 and fp32 gradients agree in the upper third of the language model (cosine
similarity 0.96 or above) but are nearly orthogonal in the lower third. On image prompts, we
trace the effect to an ill-conditioned backward path. The fp32 gradient is unchanged by element-wise relative noise of
10\textsuperscript{−6} on hidden states, confirming that fp32 is a valid reference. At 10\textsuperscript{−4}, however, its cosine similarity with the
noise-free gradient falls below 0.9 in 7 of 10 measurements, whereas bf16 forward states differ from fp32 states by
about 1\% per token. State swaps and backward-only interventions localize this sensitivity in Gemma-4-12B to the MLPs of
layers 19--21 at image-token positions, where the gradient can be dominated by a single low-information patch. Detaching
that path in the backward pass (while leaving the forward pass unchanged) keeps the gradient stable at 10\textsuperscript{−4} (cosine
similarity 0.97 or above across all five measurements), lowers the median pre-clipping gradient norm from 8.5 to 0.63, and improves
validation loss on a CT lesion-grounding task from 0.336 ± 0.019 to 0.296 ± 0.003 (four seeds; every modified run beats
every baseline run). The gain disappears when the lower layers are frozen, and an fp32 run still exhibits gradient
spikes (maximum norm 39,862) and fails to match this improvement. These training results cover one task and one model.
\end{abstract}

\section{Introduction}\label{introduction}

bf16 is the default precision for fine-tuning large language models, and suspected precision problems are typically
debugged by recomputing in fp32 {[}1{]}. When the two disagree, the standard remedy is to pay for higher precision. We
describe a case where it is better to repair the computation that makes precision matter.

While validating a learner for reinforcement-learning fine-tuning of Gemma-4 on CT images, we found that the bf16
gradient in the lower language-model layers was uncorrelated with the fp32 gradient. We report three findings.

\begin{itemize}
\tightlist
\item
  \textbf{Phenomenon (§3).} On long prompts, the first-step bf16 and fp32 LoRA gradients in the lower third of Gemma-4 12B
  and 31B have a cosine similarity near 0. This holds across CT images, natural photos, and long text, with both
  full-precision and 4-bit weights. Short text prompts and Qwen3.8-27B are affected far less.
\item
  \textbf{Diagnosis (§3--4).} fp32 is a valid reference, but on image prompts, element-wise noise of 10\textsuperscript{−4}---far smaller than
  the bf16--fp32 difference in forward states---leaves only 3 of 10 gradients with a cosine similarity above 0.9 relative
  to their noise-free value. In Gemma-4-12B, this sensitivity is concentrated in the MLPs of layers 19--21 at
  image-token positions.
\item
  \textbf{Consequence and fix (§5).} Standard bf16 training and training with the lower layers frozen reach similar mean
  validation loss, though the Standard runs show much greater variation across seeds. Detaching the image-token MLP
  output of layers 19--21 in the backward pass improves validation loss and task accuracy in every seed at no runtime
  cost, and does so only when the lower layers are trainable. An fp32 run failed to match this improvement.
\end{itemize}

\section{Setup}\label{setup}

\textbf{Models.} We evaluate Gemma-4 12B (48 language-model layers), 31B (60), E4B (42) {[}2{]}, and Qwen3.8-27B (64) {[}3{]}.
Gemma-4-12B has no separate vision encoder: image patches are embedded and fed directly to the language model. The
other three include a separate vision encoder, yet 31B exhibits the same effect, ruling out the encoder-free design as
the cause.

\textbf{Gradients.} We apply LoRA {[}4{]} (rank 16, α = 32) to every attention and MLP projection of the language model. Because
LoRA's B matrices start at zero, only B receives a gradient at the first step; we therefore compare ∂L/∂B for the
teacher-forced negative log-likelihood (NLL) of the answer. Agreement is measured as the cosine similarity computed in
float64 over all B matrices in a block of layers. Figure 1 uses six blocks of equal depth; subsequent experiments on
12B use layers 0--11 as the ``lowest block''.

\textbf{Prompts.} We use CT lesion-grounding prompts of 2,400--3,600 tokens, of which two or three CT images account for
2,100--3,200 tokens. Control prompts replace the images with natural photos {[}5{]} or omit them altogether, yielding short
text prompts (200--350 tokens) unless padded with unrelated natural text.

\textbf{Precision.} In the ``bf16'' setting, both weights and arithmetic run in bf16. In ``fp32'', the same weights are cast to
fp32 (an exact conversion), and attention uses PyTorch's deterministic ``math'' kernel. Repeated runs yield consistent
gradients (cosine ≥ 0.9989), and an fp32 backward pass on bf16 forward states reproduces the bf16 gradient
(cosine ≥ 0.996). The disagreement therefore arises from the forward states rather than from rounding in the backward
pass (Appendix A).

\section{The lower-layer bf16 gradient is unreliable on long prompts}\label{the-lower-layer-bf16-gradient-is-unreliable-on-long-prompts}

\begin{figure}[t]
\centering
\includegraphics[width=1\linewidth,height=\textheight,keepaspectratio]{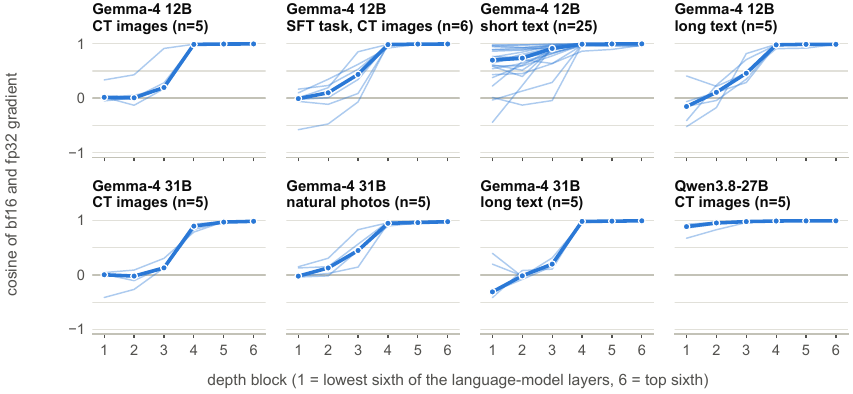}
\caption{\textbf{Agreement between bf16 and fp32 LoRA gradients by depth.} Thin lines represent individual prompts and
bold lines show medians; each block spans one-sixth of the language-model layers. Short text prompts contain 200--350
tokens, and long text prompts about 2,200. The Gemma-4-12B panels use full-precision weights; the others use 4-bit NF4
weights {[}6{]} with bf16 or fp32 arithmetic (Appendix A). ``SFT task'' uses the training prompts from
§5.}
\end{figure}

\textbf{bf16 and fp32 disagree in the lower layers.} On CT images, the two lowest blocks of Gemma-4-12B have median cosines
of 0.02 and 0.01, whereas blocks 4--6 remain at 0.98 or above (Figure 1). Gemma-4-31B behaves similarly on CT images
(lowest block −0.41 to +0.04) and natural photos (−0.04 to +0.15), showing that the effect is not specific to medical
images. On the same CT prompts, Qwen3.8-27B stays at 0.67--0.93, while Gemma-4-E4B falls in between (0.08--0.96).

\textbf{Length matters, not just images.} Short text-only prompts show milder disagreement (lowest-block median around 0.7),
but they are ten times shorter than the image prompts. When padded with natural text to about 2,260 tokens, text-only
prompts yield bf16--fp32 cosines of −0.47 to +0.36 in the lowest 12 layers of 12B (median −0.13)---as severe as with
images---whereas at about 730 tokens the median is 0.48. Gemma-4-31B yields −0.41 to +0.40 at 2,170 tokens, and fares no
better when padded with a single repeated sentence. A long context alone is sufficient.

\textbf{How much perturbation the gradient tolerates.} We multiplied every element of every layer's output in the fp32
forward pass by 1 + σε, with standard Gaussian noise ε, and compared the resulting lowest-block gradient with the
unperturbed gradient on five image prompts in Gemma-4-12B (Figure 2a). Up to σ = 10\textsuperscript{−6}, nothing changes: all 25 runs across 10\textsuperscript{−7}, 3×10\textsuperscript{−7}, and 10\textsuperscript{−6}
maintain a cosine of at least 0.99, confirming that fp32 (whose rounding error is near 10\textsuperscript{−7}) is a valid reference.
At σ = 10\textsuperscript{−5}, 2 of 10 runs fall below 0.9; at 10\textsuperscript{−4}, only 3 of 10 remain above 0.9; and at 10\textsuperscript{−3}, only one does
while half turn negative. The bf16 forward states differ from fp32 states by a median of 0.3--1.3\% per token
(measured as the per-token relative L2 difference, taking the median across tokens and reporting the range across
layers; a single bf16 rounding step introduces at most 4×10\textsuperscript{−3} relative error). Because this discrepancy is 30--130 times the noise
that already breaks most of these gradients, the bf16 gradient in these layers carries almost no information about the fp32 gradient. Long text
prompts are less fragile (0.91--0.99 at 10\textsuperscript{−4}), though bf16 still breaks them.

\begin{figure}[t]
\centering
\includegraphics[width=1\linewidth,height=\textheight,keepaspectratio]{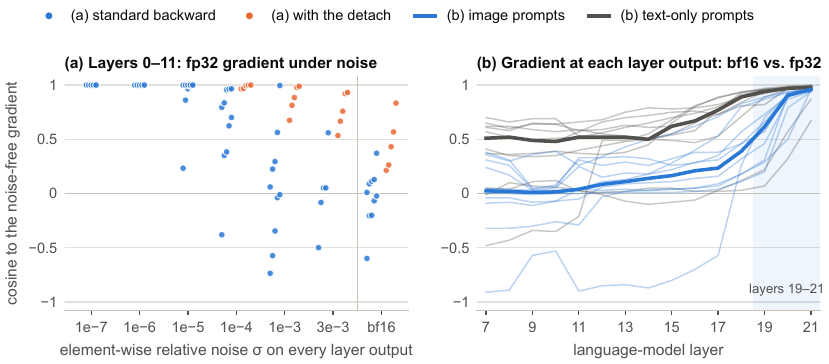}
\caption{\textbf{(a)} Cosine similarity between the lowest-block fp32 gradient with and without element-wise relative
noise σ applied to every layer output (five image prompts, Gemma-4-12B). Each point is compared against its own
noise-free gradient (standard or detached). The final column compares bf16 with fp32; ``Detach'' refers to the
intervention in §4--5. \textbf{(b)} Cosine similarity between bf16 and fp32 gradients at each layer's output (10 image
prompts and 11 short text-only prompts).}
\end{figure}

\section{Localization: the MLP on image tokens in layers 19--21}\label{localization-the-mlp-on-image-tokens-in-layers-1921}

\textbf{Where the gradients diverge.} Tracing backward from the loss, bf16 and fp32 gradients at each layer's output agree
down to layer 21 (median cosine 0.96 on image prompts) before collapsing to 0.6 at layer 19 and 0.2 at layer 17
(Figure 2b). Because these are gradients with respect to hidden states, the issue originates in the base model's
backward pass rather than in LoRA. Long text prompts behave differently: agreement degrades gradually, and in two of
five prompts it is already low at layer 21 (Appendix F). The localization below is specific to image prompts.

\textbf{Which states and which component.} To isolate the source of sensitivity, we applied the backward map J\textsuperscript{T} from layer
21 to layer 11 to a fixed vector g (the fp32 gradient at layer 21), comparing its evaluation on all-fp32 states against
runs where selected layers used bf16 states (Table 1). Substituting bf16 states in layers 19--21 alone reproduces the
full collapse, whereas doing so in layers 12--18 has little effect. While the standard backward map amplifies g by
11--206×, detaching the MLP output at image-token positions in layers 19--21 (in the backward pass only) restores
agreement to 0.94 and moderates the gain to roughly 7. Detaching only the gate activation recovers most of this
agreement (0.86), but detaching the up-projection offers little recovery (0.27, with gain unchanged).

\begin{longtable}[]{@{}
  >{\raggedright\arraybackslash}p{(\linewidth - 6\tabcolsep) * \real{0.2500}}
  >{\raggedright\arraybackslash}p{(\linewidth - 6\tabcolsep) * \real{0.2500}}
  >{\raggedright\arraybackslash}p{(\linewidth - 6\tabcolsep) * \real{0.2500}}
  >{\raggedright\arraybackslash}p{(\linewidth - 6\tabcolsep) * \real{0.2500}}@{}}
\caption{\textbf{Sensitivity of the backward map from layer 21 to layer 11 to bf16 forward states} (Gemma-4-12B,
5 image prompts; median {[}min, max{]}). Agreement is the cosine similarity between J\textsuperscript{T}g evaluated at fp32 states and at
states where the listed layers use bf16 values. Gain is the ratio of the norm of J\textsuperscript{T}g to the norm of g at fp32 states.
Full table in Appendix F.}\tabularnewline
\toprule\noalign{}
\begin{minipage}[b]{\linewidth}\raggedright
Backward pass
\end{minipage} & \begin{minipage}[b]{\linewidth}\raggedright
bf16 states in
\end{minipage} & \begin{minipage}[b]{\linewidth}\raggedright
Agreement
\end{minipage} & \begin{minipage}[b]{\linewidth}\raggedright
Gain
\end{minipage} \\
\midrule\noalign{}
\endfirsthead
\toprule\noalign{}
\begin{minipage}[b]{\linewidth}\raggedright
Backward pass
\end{minipage} & \begin{minipage}[b]{\linewidth}\raggedright
bf16 states in
\end{minipage} & \begin{minipage}[b]{\linewidth}\raggedright
Agreement
\end{minipage} & \begin{minipage}[b]{\linewidth}\raggedright
Gain
\end{minipage} \\
\midrule\noalign{}
\endhead
\bottomrule\noalign{}
\endlastfoot
Standard & layers 12--14 or 15--18 & 0.99 and 0.93 (medians) & 21 {[}11, 206{]} \\
Standard & layers 19--21 & 0.07 {[}−0.24, 0.75{]} & 21 {[}11, 206{]} \\
MLP output detached, image tokens, L19--21 & all layers & \textbf{0.94} {[}0.69, 0.98{]} & 6.8 {[}4.8, 8.9{]} \\
Gate activation detached, image tokens, L19--21 & all layers & 0.86 {[}0.56, 0.97{]} & 7.1 {[}6.0, 9.1{]} \\
Up-projection detached, image tokens, L19--21 & all layers & 0.27 {[}0.07, 0.86{]} & 21 {[}10, 145{]} \\
\end{longtable}

\textbf{A few low-information patches dominate.} On the prompt with the largest lower-layer gradient, 54\% of the squared
norm of the gradient arriving at layer 11 concentrates on a single image token---not on the lesion, but on a dark,
low-contrast patch near the image border. After detaching the MLP path in layers 19--21, the largest single-token share
drops to 8\% (on a text token). On two other prompts, the top token is a blank patch with over 100 identical copies in
the image (Appendix F). Five prompts are too few for a claim, but this pattern resembles the high-norm
background tokens observed in vision transformers {[}17{]}.

\textbf{The path carries little forward signal.} Zeroing these same MLP outputs in the base model's forward pass raises the
§5 validation NLL only slightly, from 0.653 to 0.697---compared with 1.33 when zeroing text-token outputs in the same
layers, and 1.27 when zeroing image-token outputs in layers 4--6 (Appendix G). Other mid-depth layers are similarly
dispensable at image tokens, so layers 19--21 are unremarkable in their forward contribution; what sets them apart is
how much gradient they transmit backward.

\textbf{Detaching improves gradient conditioning.} With this intervention, the full lowest-block fp32 gradient remains
unaffected by noise of σ = 10\textsuperscript{−4} (cosine ≥ 0.97) and retains a cosine of 0.54--0.93 even at σ = 3×10\textsuperscript{−3} (Figure 2a).
Agreement between bf16 and fp32 in layers 0--11 rises to a median cosine of 0.43, up from −0.02---improved, though not fully resolved
(Appendix D).

\section{Does training care?}\label{does-training-care}

We trained Gemma-4-12B with supervised fine-tuning (SFT) in bf16 on a private CT lesion-grounding task: given a key slice with a marked liver lesion
and a second view with voxel tick labels, the model outputs the lesion's center and bounding box. Training ran for 300
steps with a batch size of 8 examples using AdamW {[}7{]} and gradient clipping at norm 1 (Appendix A). All arms share the
same data order. Four arms are evaluated across four LoRA initialization seeds each: \textbf{Standard}; \textbf{Detach}, which
replaces the image-token MLP output in layers 19--21 with its stop-gradient during training; \textbf{Frozen}, which omits
LoRA updates in layers 0--18; and \textbf{Detach + Frozen}. One additional run trains the Standard configuration in fp32.

\begin{figure}[t]
\centering
\includegraphics[width=1\linewidth,height=\textheight,keepaspectratio]{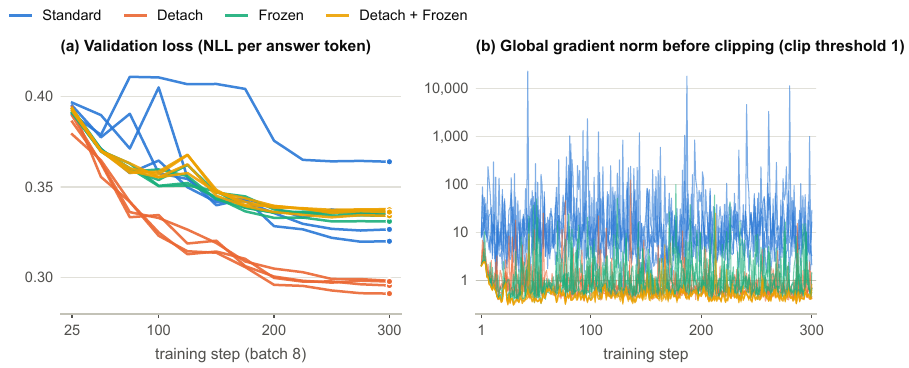}
\caption{\textbf{Training Gemma-4-12B on CT lesion grounding.} \textbf{(a)} Validation NLL on 80 held-out lesions from 80
scans not seen during training; every run starts at 0.653. \textbf{(b)} Global gradient norm before clipping (log
scale).}
\end{figure}

\begin{longtable}[]{@{}
  >{\raggedright\arraybackslash}p{(\linewidth - 10\tabcolsep) * \real{0.1667}}
  >{\raggedright\arraybackslash}p{(\linewidth - 10\tabcolsep) * \real{0.1667}}
  >{\raggedright\arraybackslash}p{(\linewidth - 10\tabcolsep) * \real{0.1667}}
  >{\raggedright\arraybackslash}p{(\linewidth - 10\tabcolsep) * \real{0.1667}}
  >{\raggedright\arraybackslash}p{(\linewidth - 10\tabcolsep) * \real{0.1667}}
  >{\raggedright\arraybackslash}p{(\linewidth - 10\tabcolsep) * \real{0.1667}}@{}}
\caption{\textbf{Results after 300 steps} (mean ± s.d. over four seeds; single seed for fp32). ``In box'' denotes the
fraction of predictions whose center falls inside the ground-truth box, evaluated with greedy decoding on 80 validation
examples.}\tabularnewline
\toprule\noalign{}
\begin{minipage}[b]{\linewidth}\raggedright
Arm
\end{minipage} & \begin{minipage}[b]{\linewidth}\raggedright
Val. NLL
\end{minipage} & \begin{minipage}[b]{\linewidth}\raggedright
Center error (voxels)
\end{minipage} & \begin{minipage}[b]{\linewidth}\raggedright
In box
\end{minipage} & \begin{minipage}[b]{\linewidth}\raggedright
Median grad. norm
\end{minipage} & \begin{minipage}[b]{\linewidth}\raggedright
Max grad. norm
\end{minipage} \\
\midrule\noalign{}
\endfirsthead
\toprule\noalign{}
\begin{minipage}[b]{\linewidth}\raggedright
Arm
\end{minipage} & \begin{minipage}[b]{\linewidth}\raggedright
Val. NLL
\end{minipage} & \begin{minipage}[b]{\linewidth}\raggedright
Center error (voxels)
\end{minipage} & \begin{minipage}[b]{\linewidth}\raggedright
In box
\end{minipage} & \begin{minipage}[b]{\linewidth}\raggedright
Median grad. norm
\end{minipage} & \begin{minipage}[b]{\linewidth}\raggedright
Max grad. norm
\end{minipage} \\
\midrule\noalign{}
\endhead
\bottomrule\noalign{}
\endlastfoot
Standard & 0.336 ± 0.019 & 16.3 ± 2.1 & 75\% ± 10 & 8.5 & 23,307 \\
Frozen below layer 19 & 0.334 ± 0.003 & 16.0 ± 0.4 & 74\% ± 4 & 0.73 & 100 \\
Detach + Frozen & 0.336 ± 0.002 & 17.0 ± 0.5 & 71\% ± 3 & 0.46 & 25 \\
Standard, fp32 & 0.332 & 16.2 & 81\% & 7.7 & 39,862 \\
\textbf{Detach} & \textbf{0.296 ± 0.003} & \textbf{10.8 ± 0.5} & \textbf{88\% ± 2} & 0.63 & 125 \\
\end{longtable}

\textbf{Detaching helps in every seed.} Every Detach run outperforms all runs from every other arm on validation NLL, center
error, and in-box rate (Table 2, Figure 3a). Within-seed validation NLL improvements over Standard are 0.066, 0.029,
0.031, and 0.037; a paired sign-flip test yields a one-sided p = 1/16, the lowest value achievable with four pairs.
Given four seeds, a single data order, and a test selected after the first three seeds had finished, these statistics should be
treated as exploratory (Appendix A).

\textbf{Standard and Frozen reach similar means.} Their mean validation NLLs are 0.336 and 0.334---which does not establish
equivalence---while seed-to-seed variation is six times larger for Standard.

\textbf{The gain requires trainable lower layers.} Detach + Frozen applies the stop-gradient intervention while updating
only layers 19 and above. Across all four seeds, detaching improved validation NLL by 0.029--0.066 when layers 0--18 were
trainable, but changed it by only −0.001 to +0.003 when those layers were frozen. These arms do not separate whether
reduced clipping or altered updates inside layers 19--21 also contribute when the lower layers train.

\textbf{fp32 falls short of the fix.} On seed 0, Standard reaches 0.364 in bf16 and 0.332 in fp32, whereas Detach reaches
0.298 in bf16. fp32 improves over its matched bf16 baseline but remains well above Detach in loss---and still suffers
from the same clipped, spiky gradients (median norm 7.7, maximum 39,862) at twice the memory footprint. This
comparison rests on a single seed.

\textbf{Gradient norms.} In Standard runs, 99.7\% of steps trigger gradient clipping and the pre-clipping norm reaches 23,307
(Figure 3b); under Detach, only 18\% of steps are clipped and the maximum norm drops to 125. The gradient propagating to
the lower layers through this path inflates the global norm roughly tenfold and drives the spikes. Per-layer clipping
would shield the upper layers, much as Frozen does, but would still update the lower layers along the ill-conditioned
direction. Step time is unchanged.

\section{Related work}\label{related-work}

Mixed-precision training {[}1{]} treats fp32 as the reference. Rounding errors at inference have been reported to either
vanish or amplify chaotically across LLM layers {[}8{]}, without examining gradients; we quantify gradient perturbation
tolerance directly and show that bf16 errors exceed it. Signal-propagation theory links exploding backward gain to
chaotic forward dynamics at initialization {[}9, 10{]}. Although gradient clipping {[}11{]} masks the norm spikes here, it still
leaves the affected layers receiving unusable updates. A community issue {[}12{]} similarly reports gradient-norm spikes
during text-only LoRA fine-tuning of Gemma-4-31B, which the reporter traced to attention in three mid-depth layers; a
proposed explanation in the thread has not been confirmed by the maintainers. Our state-swap localization follows
activation patching {[}13{]}. Selective stop-gradients are widely used in self-supervised learning {[}14{]} and have been
applied in vision-language models to prevent gradient interference from the language objective {[}15{]}. Bitwise alignment
of sampler and trainer log-probabilities {[}16{]} targets train--inference consistency, which is distinct from
state-sensitive backward conditioning.

\section{Limitations and takeaway}\label{limitations-and-takeaway}

\begin{itemize}
\tightlist
\item
  \textbf{Training evidence is narrow:} one task, one model, 300 steps, one data order, and a single fp32 seed.
\item
  \textbf{The mechanism and fix were tested only on Gemma-4-12B image prompts,} using 5--10 prompts and first-step gradients.
  Long text prompts exhibit a different depth profile across five prompts (Appendix F); because we did not run
  state-swap or detach experiments on them, their mechanism remains unlocalized and unaddressed here. Why Qwen3.8-27B
  is largely spared also remains unexplained, and our cross-model probes used 4-bit weights.
\item
  \textbf{The detached gradient is biased} (cosine 0.01--0.35 with the standard fp32 gradient). We show that it trains
  better, not that it better estimates the true gradient.
\end{itemize}

When bf16 and fp32 gradients disagree, measure how much perturbation the fp32 gradient tolerates before paying for
higher precision. If the gradient breaks under perturbations far smaller than the bf16--fp32 state difference, the
backward path is ill-conditioned, and repairing it can help more than adding bits. Test early whether training cares:
a short multi-arm training run settled what gradient comparisons alone could not.

\section*{References}\label{references}
\addcontentsline{toc}{section}{References}

\begin{enumerate}
\def\labelenumi{\arabic{enumi}.}
\tightlist
\item
  P. Micikevicius et al.~Mixed precision training. \emph{ICLR}, 2018.
\item
  Gemma Team. Gemma 4 technical report. arXiv:2607.02770, 2026.
\item
  Qwen Team. Qwen3.8-Max: A new bar for coding and cowork. Blog post and model card (\url{https://huggingface.co/Qwen/Qwen3.8-27B}), August 2026.
\item
  E. Hu et al.~LoRA: Low-rank adaptation of large language models. \emph{ICLR}, 2022.
\item
  xAI. RealWorldQA. Dataset released with the Grok-1.5 Vision preview, 2024. \url{https://huggingface.co/datasets/xai-org/RealworldQA}
\item
  T. Dettmers, A. Pagnoni, A. Holtzman, L. Zettlemoyer. QLoRA: Efficient finetuning of quantized LLMs. \emph{NeurIPS}, 2023.
\item
  I. Loshchilov, F. Hutter. Decoupled weight decay regularization. \emph{ICLR}, 2019.
\item
  C. M. Islam, A. Villarreal, M. Nishino, S. Salman, X. Liu. Numerical instability and chaos: Quantifying the unpredictability of large language models. arXiv:2604.13206, 2026.
\item
  B. Poole, S. Lahiri, M. Raghu, J. Sohl-Dickstein, S. Ganguli. Exponential expressivity in deep neural networks through transient chaos. \emph{NeurIPS}, 2016.
\item
  J. Pennington, S. Schoenholz, S. Ganguli. Resurrecting the sigmoid in deep learning through dynamical isometry: theory and practice. \emph{NeurIPS}, 2017.
\item
  R. Pascanu, T. Mikolov, Y. Bengio. On the difficulty of training recurrent neural networks. \emph{ICML}, 2013.
\item
  Gemma 4: Exploding pre-clip gradient norms during LoRA fine-tuning of gemma-4-31B-it. GitHub issue \#45676, huggingface/transformers, April 2026.
\item
  K. Meng, D. Bau, A. Andonian, Y. Belinkov. Locating and editing factual associations in GPT. \emph{NeurIPS}, 2022.
\item
  X. Chen, K. He. Exploring simple Siamese representation learning. \emph{CVPR}, 2021.
\item
  Z. Yuan, R. Yao, C. Zheng, Y. Zhao, D. Dong, M. Zhang. Detached skip-links and R-probe: Decoupling feature aggregation from gradient propagation for MLLM OCR. \emph{ICML}, 2026. arXiv:2603.20020.
\item
  Y. Zhang, L. Chai, L. Wang, T. Bao, W. Yu. Unlocking stable on-policy RL on TPU with bitwise-zero train--inference mismatch. Google Developers Blog, 16 November 2026.
\item
  T. Darcet, M. Oquab, J. Mairal, P. Bojanowski. Vision transformers need registers. \emph{ICLR}, 2024. arXiv:2309.16588.
\end{enumerate}

\appendix

\section{Implementation details}\label{implementation-details}

\textbf{The detach.} The fix is a patch to the MLP module's forward function. It changes no forward value.

\begin{verbatim}
# Backward-only detach of the MLP output at image-token positions.
# img_mask: bool tensor [seq_len], True at the image tokens of the current example.
DETACH_LAYERS = {19, 20, 21}

def mlp_forward(self, x, _orig=TextMLP.forward):
    y = _orig(self, x)                       # W_down(act(W_gate x) * W_up x)
    if layer_index[id(self)] in DETACH_LAYERS:
        y = torch.where(img_mask[..., None], y.detach(), y)
    return y

TextMLP.forward = mlp_forward    # patch both Gemma4TextMLP and Gemma4UnifiedTextMLP
\end{verbatim}

With the output detached at image tokens, no gradient passes through these three MLPs at those positions, neither to
the layers below nor to the LoRA matrices inside the MLPs. Gradient still flows through the residual stream, through
attention, and through the same MLPs at text positions.

\textbf{Controls} (12B; 10 image and 11 short text prompts). A repeated fp32 gradient is identical, and a repeated bf16
gradient has cosine ≥ 0.9989. An fp32 backward pass run on bf16 forward states reproduces the bf16 gradient with
cosine ≥ 0.996. The bf16 and fp32 forward states differ by a median of 0.3--1.3\% per token. fp32 with PyTorch's default
attention kernel agrees with the math kernel at 0.99 or more on five image prompts.

\textbf{Weights in Figure 1.} The four Gemma-4-12B panels use full-precision weights and the math attention kernel for
fp32. The other four use 4-bit NF4 weights with bf16 or fp32 as the compute precision on the same dequantized
weights, and PyTorch's default attention kernel; the 31B model does not fit on our GPU in fp32 with full-precision
weights. The long-text panels pad text-only prompts with natural text. Gemma-4-E4B, not shown, has a lowest-block
cosine of 0.08--0.96 on CT prompts. The vision encoders have 27 layers in 31B and Qwen3.8-27B and 16 in
E4B. Six prompts from the supervised task give a lowest-block cosine of −0.58 to +0.17 on 12B. Short text-only
prompts have a lowest-block median of about 0.7 on both 12B and 31B; 3 of 25 on 12B fall below 0.1.

\textbf{Interventions in Table 1.} With MLP(x) = W\textsubscript{down}(g ⊙ u), g = act(W\textsubscript{gate}x) and u = W\textsubscript{up}x, ``MLP output detached''
stops the gradient at MLP(x), ``gate activation detached'' at g, and ``up-projection detached'' at u, each only at the
listed token positions and layers.

\textbf{Noise.} In the element-wise test of §3, each layer's output h (all layers but the last) is replaced by
h + σ·h·ε, with ε standard Gaussian noise drawn from a fixed seed, so every element moves in proportion to its own
size, as rounding does. fp32 has a relative rounding step of about 1.2×10\textsuperscript{−7} and bf16 of about 7.8×10\textsuperscript{−3}, so the largest rounding error is
half of that (3.9×10\textsuperscript{−3} for bf16). An
earlier version of the test (Appendix B) used h + σ·r·ε, where r is the root-mean-square of that token's state.

\textbf{State substitution.} We store the input of every layer from a plain bf16 forward pass and from a plain fp32 one.
In the test pass, which runs in fp32 arithmetic, the value entering each layer is overwritten before the layer runs:
\texttt{h\ =\ stored\ +\ (h\ -\ h.detach())}, with \texttt{stored} taken from the bf16 pass for layers in the chosen window and from the
fp32 pass for all others. Each layer is therefore recomputed from its substituted input, so its local Jacobian is
evaluated at that input, and the graph stays connected through the unchanged \texttt{h}. ``bf16 states in layers 19--21''
means layers 19, 20 and 21 receive bf16 inputs and every other layer receives fp32 inputs. In Table 1 the vector g
is fixed, the map runs from the output of layer 21 back to the output of layer 11, and the same substitution is
applied with and without each detach.

\textbf{Training.} LoRA rank 16, α = 32, no dropout, 65.6M trainable parameters. AdamW with β = (0.9, 0.99) and no weight
decay; learning rate 10\textsuperscript{−4} with 15 warm-up steps and cosine decay; batch 8 by accumulating per-example gradients;
clipping at global norm 1; bf16 weights and arithmetic; gradient checkpointing; up to 1,120 tokens per image. The
dataset has 14,870 training examples, of which the 300 steps use 2,400, and 200 validation examples, of which we
evaluate the same 80 in every run: the first 80 of the even-indexed examples, which are 80 lesions from 80 scans.
No validation scan is in the training set; we did not check whether any patient contributes scans to both.
Center error is the Euclidean distance between predicted and true centers in voxel indices, without scaling by voxel
spacing. Validation NLL is computed every 25 steps. All runs used one NVIDIA RTX PRO 6000 Blackwell GPU (96 GB) with
PyTorch defaults, TF32 disabled in training, and the math attention kernel for the fp32 references. The fp32
training run casts the bf16 weights to fp32 and trains with fp32 arithmetic and PyTorch's default attention kernel;
it needs about 49 GB against 25 GB for bf16. The first statistical test (an unpaired exact
permutation test on final validation NLL, one-sided p = 1/70 for four against four runs) was fixed after the first
three of the eight Standard and Detach runs had finished and before the other five; the paired test of §5 was added
later.

\section{Noise scaled by the token's RMS}\label{noise-scaled-by-the-tokens-rms}

Our first noise test scaled the noise by each token's root-mean-square instead of by each element. This perturbs
small elements by far more than their own rounding, so it is not a model of fp32 rounding, and we do not use it for
the claims of §3. We keep it because it shows the sign flip of §4: at σ = 10\textsuperscript{−7} it breaks the gradient on image
prompts (Table 3), where element-wise noise of the same σ changes nothing (cosine 0.99 or more in 20 runs). We think
the difference comes from a few very large channels inflating the RMS, but we did not test this. Short text-only
prompts are stable under the RMS-scaled noise at 10\textsuperscript{−7} and 3×10\textsuperscript{−7} (33 runs at 0.95 or more).

\begin{longtable}[]{@{}
  >{\raggedright\arraybackslash}p{(\linewidth - 12\tabcolsep) * \real{0.1429}}
  >{\raggedright\arraybackslash}p{(\linewidth - 12\tabcolsep) * \real{0.1429}}
  >{\raggedright\arraybackslash}p{(\linewidth - 12\tabcolsep) * \real{0.1429}}
  >{\raggedright\arraybackslash}p{(\linewidth - 12\tabcolsep) * \real{0.1429}}
  >{\raggedright\arraybackslash}p{(\linewidth - 12\tabcolsep) * \real{0.1429}}
  >{\raggedright\arraybackslash}p{(\linewidth - 12\tabcolsep) * \real{0.1429}}
  >{\raggedright\arraybackslash}p{(\linewidth - 12\tabcolsep) * \real{0.1429}}@{}}
\caption{\textbf{Cosine between the lowest-block (layers 0--11) gradient under RMS-scaled noise and the plain fp32
gradient}, Gemma-4-12B, image prompts. Two noise seeds for σ = 10\textsuperscript{−7} and 3×10\textsuperscript{−7}. The bf16 column compares bf16
with fp32.}\tabularnewline
\toprule\noalign{}
\begin{minipage}[b]{\linewidth}\raggedright
Image prompt
\end{minipage} & \begin{minipage}[b]{\linewidth}\raggedright
bf16
\end{minipage} & \begin{minipage}[b]{\linewidth}\raggedright
σ = 10\textsuperscript{−7}, seed 1
\end{minipage} & \begin{minipage}[b]{\linewidth}\raggedright
σ = 10\textsuperscript{−7}, seed 2
\end{minipage} & \begin{minipage}[b]{\linewidth}\raggedright
σ = 3×10\textsuperscript{−7}, seed 1
\end{minipage} & \begin{minipage}[b]{\linewidth}\raggedright
σ = 3×10\textsuperscript{−7}, seed 2
\end{minipage} & \begin{minipage}[b]{\linewidth}\raggedright
σ = 10\textsuperscript{−5}
\end{minipage} \\
\midrule\noalign{}
\endfirsthead
\toprule\noalign{}
\begin{minipage}[b]{\linewidth}\raggedright
Image prompt
\end{minipage} & \begin{minipage}[b]{\linewidth}\raggedright
bf16
\end{minipage} & \begin{minipage}[b]{\linewidth}\raggedright
σ = 10\textsuperscript{−7}, seed 1
\end{minipage} & \begin{minipage}[b]{\linewidth}\raggedright
σ = 10\textsuperscript{−7}, seed 2
\end{minipage} & \begin{minipage}[b]{\linewidth}\raggedright
σ = 3×10\textsuperscript{−7}, seed 1
\end{minipage} & \begin{minipage}[b]{\linewidth}\raggedright
σ = 3×10\textsuperscript{−7}, seed 2
\end{minipage} & \begin{minipage}[b]{\linewidth}\raggedright
σ = 10\textsuperscript{−5}
\end{minipage} \\
\midrule\noalign{}
\endhead
\bottomrule\noalign{}
\endlastfoot
Prompt 1 & −0.02 & −0.99 & −0.97 & −0.66 & +0.99 & −0.60 \\
Prompt 2 & +0.37 & +1.00 & +1.00 & +0.05 & +0.07 & +0.34 \\
Prompt 3 & +0.01 & +0.91 & +0.99 & +0.65 & +0.92 & +0.19 \\
Prompt 4 & −0.07 & +0.92 & −0.41 & +0.23 & −0.41 & −0.20 \\
Prompt 5 & −0.20 & +0.86 & +0.83 & +0.88 & +0.92 & +0.47 \\
\end{longtable}

\section{Batch averaging and finite differences}\label{batch-averaging-and-finite-differences}

\textbf{Batch averaging.} We normalized each prompt's lowest-block gradient to unit length and summed over prompts. For
image prompts, the sum over five prompts has cosine 0.53 with the same sum under RMS-scaled σ = 10\textsuperscript{−7} noise
(Appendix B) and 0.20 under σ = 3×10\textsuperscript{−7}; the bf16 and fp32 sums over ten prompts have cosine 0.08. For text-only prompts the fp32 sum is
unchanged by σ = 3×10\textsuperscript{−7} noise (cosine 1.00), and the bf16 and fp32 sums have cosine 0.29. Gradients of different
prompts are almost orthogonal (mean cosine 0.003 for images, 0.018 for text), so signal and noise grow together with
batch size at this scale.

\textbf{Finite differences.} Starting from B = 0, we moved the lowest-block B matrices by a step of L2 length η against
the unit gradient direction and divided the actual change in the fp32 loss by the first-order prediction, −η times
the gradient norm (Table 4). A ratio near 1 means the gradient predicts the loss. For the fp32 gradient the ratio is 0.33--1.18 at
η = 10\textsuperscript{−5}, and by η = 10\textsuperscript{−4} it is 0.18 or less on three of five prompts, one of them negative. The bf16 direction
gives a ratio of 0.10 or less, often negative, on four of five prompts at η = 10\textsuperscript{−5} to 10\textsuperscript{−4}. So even the fp32
gradient describes the loss only within a very small radius. Appendix D gives the same test for the detach of §5.

\begin{longtable}[]{@{}
  >{\raggedright\arraybackslash}p{(\linewidth - 14\tabcolsep) * \real{0.1250}}
  >{\raggedright\arraybackslash}p{(\linewidth - 14\tabcolsep) * \real{0.1250}}
  >{\raggedright\arraybackslash}p{(\linewidth - 14\tabcolsep) * \real{0.1250}}
  >{\raggedright\arraybackslash}p{(\linewidth - 14\tabcolsep) * \real{0.1250}}
  >{\raggedright\arraybackslash}p{(\linewidth - 14\tabcolsep) * \real{0.1250}}
  >{\raggedright\arraybackslash}p{(\linewidth - 14\tabcolsep) * \real{0.1250}}
  >{\raggedright\arraybackslash}p{(\linewidth - 14\tabcolsep) * \real{0.1250}}
  >{\raggedright\arraybackslash}p{(\linewidth - 14\tabcolsep) * \real{0.1250}}@{}}
\caption{\textbf{Ratio of actual to predicted loss decrease along the lowest-block gradient direction} (loss
evaluated in fp32; one row per prompt). ``Detach'' rows use the gradient computed with the detach of §5 and its own
norm for the prediction; the loss is the unmodified one.}\tabularnewline
\toprule\noalign{}
\begin{minipage}[b]{\linewidth}\raggedright
Direction
\end{minipage} & \begin{minipage}[b]{\linewidth}\raggedright
Image prompt
\end{minipage} & \begin{minipage}[b]{\linewidth}\raggedright
η = 10\textsuperscript{−6}
\end{minipage} & \begin{minipage}[b]{\linewidth}\raggedright
η = 10\textsuperscript{−5}
\end{minipage} & \begin{minipage}[b]{\linewidth}\raggedright
η = 3×10\textsuperscript{−5}
\end{minipage} & \begin{minipage}[b]{\linewidth}\raggedright
η = 10\textsuperscript{−4}
\end{minipage} & \begin{minipage}[b]{\linewidth}\raggedright
η = 3×10\textsuperscript{−4}
\end{minipage} & \begin{minipage}[b]{\linewidth}\raggedright
η = 10\textsuperscript{−3}
\end{minipage} \\
\midrule\noalign{}
\endfirsthead
\toprule\noalign{}
\begin{minipage}[b]{\linewidth}\raggedright
Direction
\end{minipage} & \begin{minipage}[b]{\linewidth}\raggedright
Image prompt
\end{minipage} & \begin{minipage}[b]{\linewidth}\raggedright
η = 10\textsuperscript{−6}
\end{minipage} & \begin{minipage}[b]{\linewidth}\raggedright
η = 10\textsuperscript{−5}
\end{minipage} & \begin{minipage}[b]{\linewidth}\raggedright
η = 3×10\textsuperscript{−5}
\end{minipage} & \begin{minipage}[b]{\linewidth}\raggedright
η = 10\textsuperscript{−4}
\end{minipage} & \begin{minipage}[b]{\linewidth}\raggedright
η = 3×10\textsuperscript{−4}
\end{minipage} & \begin{minipage}[b]{\linewidth}\raggedright
η = 10\textsuperscript{−3}
\end{minipage} \\
\midrule\noalign{}
\endhead
\bottomrule\noalign{}
\endlastfoot
fp32 & Prompt 1 & +0.68 & +0.42 & +0.20 & +0.09 & +0.01 & +0.02 \\
fp32 & Prompt 2 & +0.24 & +0.87 & +0.89 & +1.16 & +0.91 & +0.61 \\
fp32 & Prompt 3 & +0.44 & +0.84 & +0.89 & +0.77 & −0.02 & +0.44 \\
fp32 & Prompt 4 & +0.84 & +0.33 & +0.33 & −0.36 & −0.18 & −0.06 \\
fp32 & Prompt 5 & +0.65 & +1.18 & +1.20 & +0.18 & +0.45 & +0.06 \\
bf16 & Prompt 1 & +0.64 & −0.05 & −0.03 & +0.10 & +0.61 & +0.20 \\
bf16 & Prompt 2 & −0.54 & +0.34 & +0.45 & +0.83 & +0.44 & +0.56 \\
bf16 & Prompt 3 & +0.02 & +0.00 & −0.00 & +0.00 & +0.00 & +0.00 \\
bf16 & Prompt 4 & −0.04 & −0.12 & −0.08 & −0.07 & −0.04 & +0.01 \\
bf16 & Prompt 5 & −0.17 & −0.04 & −0.06 & −0.26 & −0.06 & −0.05 \\
fp32, detach & Prompt 1 & −8.13 & −0.81 & +0.97 & +6.62 & +0.90 & +1.86 \\
fp32, detach & Prompt 2 & +0.34 & +1.06 & +1.22 & +0.82 & +1.52 & +2.11 \\
fp32, detach & Prompt 3 & +0.92 & +0.98 & +0.97 & +0.96 & +0.62 & −0.93 \\
fp32, detach & Prompt 4 & +7.13 & +1.08 & +0.68 & +1.11 & +0.94 & +0.98 \\
fp32, detach & Prompt 5 & +2.73 & +2.87 & +3.79 & +0.13 & −4.30 & −2.58 \\
\end{longtable}

\section{First-step checks of the detach, and interventions that were not trained}\label{first-step-checks-of-the-detach-and-interventions-that-were-not-trained}

\textbf{Stability and agreement.} Table 5 gives the element-wise noise test of §3 per prompt, with and without the
detach of §4 (MLP output at image tokens, layers 19--21). With the detach the gradient is stable at σ = 10\textsuperscript{−4} and
degrades at 10\textsuperscript{−3}, about a factor of 10--30 later than without it. bf16 and fp32 agree better than before but far
from fully (0.21--0.83, median 0.43). The detached gradient is a different direction from the standard one: their
cosine in fp32 is 0.01--0.35. Under the RMS-scaled noise of Appendix B, the detached gradient keeps a cosine of 0.95
or more in all 20 runs at σ = 10\textsuperscript{−7} and 3×10\textsuperscript{−7}.

\begin{longtable}[]{@{}
  >{\raggedright\arraybackslash}p{(\linewidth - 10\tabcolsep) * \real{0.1667}}
  >{\raggedright\arraybackslash}p{(\linewidth - 10\tabcolsep) * \real{0.1667}}
  >{\raggedright\arraybackslash}p{(\linewidth - 10\tabcolsep) * \real{0.1667}}
  >{\raggedright\arraybackslash}p{(\linewidth - 10\tabcolsep) * \real{0.1667}}
  >{\raggedright\arraybackslash}p{(\linewidth - 10\tabcolsep) * \real{0.1667}}
  >{\raggedright\arraybackslash}p{(\linewidth - 10\tabcolsep) * \real{0.1667}}@{}}
\caption{\textbf{Lowest-block gradient under element-wise noise, with and without the detach} (Gemma-4-12B, image
prompts, same numbering as Table 3). Noise rows give the cosine between the fp32 gradient with noise and without;
two values are two noise seeds. The last rows compare bf16 with fp32 and give gradient norms in fp32.}\tabularnewline
\toprule\noalign{}
\begin{minipage}[b]{\linewidth}\raggedright
Quantity
\end{minipage} & \begin{minipage}[b]{\linewidth}\raggedright
Prompt 1
\end{minipage} & \begin{minipage}[b]{\linewidth}\raggedright
Prompt 2
\end{minipage} & \begin{minipage}[b]{\linewidth}\raggedright
Prompt 3
\end{minipage} & \begin{minipage}[b]{\linewidth}\raggedright
Prompt 4
\end{minipage} & \begin{minipage}[b]{\linewidth}\raggedright
Prompt 5
\end{minipage} \\
\midrule\noalign{}
\endfirsthead
\toprule\noalign{}
\begin{minipage}[b]{\linewidth}\raggedright
Quantity
\end{minipage} & \begin{minipage}[b]{\linewidth}\raggedright
Prompt 1
\end{minipage} & \begin{minipage}[b]{\linewidth}\raggedright
Prompt 2
\end{minipage} & \begin{minipage}[b]{\linewidth}\raggedright
Prompt 3
\end{minipage} & \begin{minipage}[b]{\linewidth}\raggedright
Prompt 4
\end{minipage} & \begin{minipage}[b]{\linewidth}\raggedright
Prompt 5
\end{minipage} \\
\midrule\noalign{}
\endhead
\bottomrule\noalign{}
\endlastfoot
Standard, σ = 10\textsuperscript{−7} & +1.00 / +1.00 & +1.00 / +1.00 & +1.00 / +1.00 & +1.00 / +1.00 & +1.00 / +1.00 \\
Standard, σ = 10\textsuperscript{−6} & +1.00 & +1.00 & +1.00 & +1.00 & +1.00 \\
Standard, σ = 10\textsuperscript{−5} & +1.00 / +1.00 & +1.00 / +1.00 & +0.99 / +1.00 & +1.00 / +0.23 & +0.86 / +0.97 \\
Standard, σ = 10\textsuperscript{−4} & +0.84 / +0.97 & +0.38 / +0.96 & +0.62 / +0.96 & +0.35 / +0.80 & +0.70 / −0.38 \\
Standard, σ = 10\textsuperscript{−3} & +0.06 / +1.00 & −0.01 / +0.56 & +0.23 / −0.04 & −0.74 / −0.35 & −0.57 / +0.30 \\
Standard, σ = 3×10\textsuperscript{−3} & +0.05 & +0.05 & +0.56 & −0.08 & −0.50 \\
Detach, σ = 10\textsuperscript{−4} & +1.00 & +0.97 & +1.00 & +1.00 & +0.97 \\
Detach, σ = 10\textsuperscript{−3} & +0.98 & +0.88 & +0.99 & +0.81 & +0.68 \\
Detach, σ = 3×10\textsuperscript{−3} & +0.93 & +0.76 & +0.92 & +0.53 & +0.67 \\
bf16 vs fp32, standard & −0.02 & +0.37 & +0.01 & −0.07 & −0.20 \\
bf16 vs fp32, detach & +0.57 & +0.26 & +0.21 & +0.83 & +0.43 \\
Detach vs standard, fp32 & +0.01 & +0.35 & +0.34 & +0.12 & +0.30 \\
Norm, standard & 73.70 & 3.38 & 4.43 & 13.72 & 11.57 \\
Norm, detach & 1.48 & 0.91 & 1.30 & 0.84 & 0.96 \\
\end{longtable}

\textbf{Finite differences.} The detach rows of Table 4 must be read differently from the others. The detached
gradient d is not the gradient of the loss, so for small steps the ratio tends to the norm of the true gradient g
divided by the norm of d, times their cosine, and not to 1. From Table 5 this limit is 1.3, 1.2 and 3.6 for prompts
2, 3 and 5, close to the measured 1.1--1.2, 1.0 and 2.9--3.8 at η = 10\textsuperscript{−5} to 3×10\textsuperscript{−5}. At η = 10\textsuperscript{−6} the loss changes
are too small to measure in fp32, also for the standard gradient. What matters is the sign at larger steps: at
η = 10\textsuperscript{−4} the loss falls along the detached direction on all five prompts, and at 3×10\textsuperscript{−4} on four, against four
and three for the standard fp32 gradient. This is modest support, on five prompts.

\textbf{Interventions that were not trained.} Two earlier candidates were tested on first-step gradients: detaching only
the gate activation at image tokens in layers 19--21, and clipping the vector--Jacobian product through those layers.
With each applied to both precisions, the lowest-block bf16 and fp32 gradients agreed no better than with the final
detach (cosines 0.06--0.47 and −0.23--0.62 over five prompts).

\section{Results per seed}\label{results-per-seed}

\begin{longtable}[]{@{}
  >{\raggedright\arraybackslash}p{(\linewidth - 14\tabcolsep) * \real{0.1250}}
  >{\raggedright\arraybackslash}p{(\linewidth - 14\tabcolsep) * \real{0.1250}}
  >{\raggedright\arraybackslash}p{(\linewidth - 14\tabcolsep) * \real{0.1250}}
  >{\raggedright\arraybackslash}p{(\linewidth - 14\tabcolsep) * \real{0.1250}}
  >{\raggedright\arraybackslash}p{(\linewidth - 14\tabcolsep) * \real{0.1250}}
  >{\raggedright\arraybackslash}p{(\linewidth - 14\tabcolsep) * \real{0.1250}}
  >{\raggedright\arraybackslash}p{(\linewidth - 14\tabcolsep) * \real{0.1250}}
  >{\raggedright\arraybackslash}p{(\linewidth - 14\tabcolsep) * \real{0.1250}}@{}}
\caption{\textbf{Final results of every run.} Seeds are LoRA initialization seeds; the data order is the same in
all runs.}\tabularnewline
\toprule\noalign{}
\begin{minipage}[b]{\linewidth}\raggedright
Arm
\end{minipage} & \begin{minipage}[b]{\linewidth}\raggedright
Seed
\end{minipage} & \begin{minipage}[b]{\linewidth}\raggedright
Val. NLL
\end{minipage} & \begin{minipage}[b]{\linewidth}\raggedright
Center error, mean
\end{minipage} & \begin{minipage}[b]{\linewidth}\raggedright
Center error, median
\end{minipage} & \begin{minipage}[b]{\linewidth}\raggedright
In box
\end{minipage} & \begin{minipage}[b]{\linewidth}\raggedright
Median grad. norm
\end{minipage} & \begin{minipage}[b]{\linewidth}\raggedright
Max grad. norm
\end{minipage} \\
\midrule\noalign{}
\endfirsthead
\toprule\noalign{}
\begin{minipage}[b]{\linewidth}\raggedright
Arm
\end{minipage} & \begin{minipage}[b]{\linewidth}\raggedright
Seed
\end{minipage} & \begin{minipage}[b]{\linewidth}\raggedright
Val. NLL
\end{minipage} & \begin{minipage}[b]{\linewidth}\raggedright
Center error, mean
\end{minipage} & \begin{minipage}[b]{\linewidth}\raggedright
Center error, median
\end{minipage} & \begin{minipage}[b]{\linewidth}\raggedright
In box
\end{minipage} & \begin{minipage}[b]{\linewidth}\raggedright
Median grad. norm
\end{minipage} & \begin{minipage}[b]{\linewidth}\raggedright
Max grad. norm
\end{minipage} \\
\midrule\noalign{}
\endhead
\bottomrule\noalign{}
\endlastfoot
Standard & 0 & 0.364 & 19.1 & 13.0 & 60\% & 7.62 & 23,307 \\
Standard & 1 & 0.320 & 14.1 & 6.7 & 84\% & 8.63 & 2,370 \\
Standard & 2 & 0.327 & 16.2 & 8.2 & 78\% & 7.61 & 1,346 \\
Standard & 3 & 0.334 & 15.6 & 8.1 & 78\% & 11.11 & 18,326 \\
Detach & 0 & 0.298 & 11.3 & 4.1 & 89\% & 0.63 & 26 \\
Detach & 1 & 0.291 & 10.2 & 4.1 & 88\% & 0.57 & 15 \\
Detach & 2 & 0.296 & 10.9 & 4.5 & 90\% & 0.67 & 46 \\
Detach & 3 & 0.298 & 10.9 & 4.1 & 86\% & 0.68 & 125 \\
Frozen & 0 & 0.331 & 15.6 & 8.2 & 79\% & 0.74 & 34 \\
Frozen & 1 & 0.335 & 15.7 & 9.8 & 70\% & 0.80 & 44 \\
Frozen & 2 & 0.334 & 16.2 & 9.1 & 76\% & 0.63 & 37 \\
Frozen & 3 & 0.337 & 16.5 & 10.0 & 72\% & 0.77 & 100 \\
Detach + Frozen & 0 & 0.334 & 16.9 & 10.8 & 68\% & 0.46 & 25 \\
Detach + Frozen & 1 & 0.338 & 16.5 & 9.2 & 74\% & 0.46 & 2 \\
Detach + Frozen & 2 & 0.337 & 16.8 & 11.7 & 71\% & 0.47 & 9 \\
Detach + Frozen & 3 & 0.336 & 17.6 & 12.0 & 70\% & 0.46 & 6 \\
Standard, fp32 & 0 & 0.332 & 16.2 & 7.3 & 81\% & 7.72 & 39,862 \\
\end{longtable}

\section{Localization details}\label{localization-details}

\begin{longtable}[]{@{}
  >{\raggedright\arraybackslash}p{(\linewidth - 6\tabcolsep) * \real{0.2500}}
  >{\raggedright\arraybackslash}p{(\linewidth - 6\tabcolsep) * \real{0.2500}}
  >{\raggedright\arraybackslash}p{(\linewidth - 6\tabcolsep) * \real{0.2500}}
  >{\raggedright\arraybackslash}p{(\linewidth - 6\tabcolsep) * \real{0.2500}}@{}}
\caption{\textbf{Full version of Table 1.}}\tabularnewline
\toprule\noalign{}
\begin{minipage}[b]{\linewidth}\raggedright
Backward pass
\end{minipage} & \begin{minipage}[b]{\linewidth}\raggedright
bf16 states in
\end{minipage} & \begin{minipage}[b]{\linewidth}\raggedright
Agreement
\end{minipage} & \begin{minipage}[b]{\linewidth}\raggedright
Gain
\end{minipage} \\
\midrule\noalign{}
\endfirsthead
\toprule\noalign{}
\begin{minipage}[b]{\linewidth}\raggedright
Backward pass
\end{minipage} & \begin{minipage}[b]{\linewidth}\raggedright
bf16 states in
\end{minipage} & \begin{minipage}[b]{\linewidth}\raggedright
Agreement
\end{minipage} & \begin{minipage}[b]{\linewidth}\raggedright
Gain
\end{minipage} \\
\midrule\noalign{}
\endhead
\bottomrule\noalign{}
\endlastfoot
Standard & layers 12--14 & 0.99 {[}0.96, 1.00{]} & 21 {[}11, 206{]} \\
Standard & layers 15--18 & 0.93 {[}0.79, 1.00{]} & 21 {[}11, 206{]} \\
Standard & layers 19--21 & 0.07 {[}−0.24, 0.75{]} & 21 {[}11, 206{]} \\
Standard & all layers & 0.16 {[}−0.25, 0.75{]} & 21 {[}11, 206{]} \\
MLP output detached, image tokens, L19--21 & all layers & 0.94 {[}0.69, 0.98{]} & 6.8 {[}4.8, 8.9{]} \\
Gate activation detached, image tokens, L19--21 & all layers & 0.86 {[}0.56, 0.97{]} & 7.1 {[}6.0, 9.1{]} \\
Gate activation detached, all tokens, L19--21 & all layers & 0.72 {[}0.05, 0.93{]} & 3.6 {[}2.8, 4.1{]} \\
Up-projection detached, image tokens, L19--21 & all layers & 0.27 {[}0.07, 0.86{]} & 21 {[}10, 145{]} \\
Gate activation detached, all tokens, L12--21 & all layers & 0.99 {[}0.98, 0.99{]} & 0.67 {[}0.65, 0.75{]} \\
\end{longtable}

We have not explained why detaching the gate on all tokens does worse than on image tokens only. The gradient at each
layer's output (Figure 2b) drops abruptly in 8 of 10 image prompts and gradually in all 11 short text prompts.

\textbf{Dominant tokens.} For each of the five image prompts we took the token with the largest share of the squared norm
of the fp32 gradient arriving at layer 11. On the prompt with the largest gradient, one image token holds 54\% and ten
hold 91\%; with the detach the largest share is 8\%, on a text token. Under the RMS-scaled noise of Appendix B this
prompt's gradient turns into its exact negative in all 82 LoRA matrices of the lowest block at once (cosine −0.99),
as expected of a gradient carried by one path. On three other prompts the top token is an image token with an 18--24\%
share, and on one it is a text token (7\%). The four image patches, located from the patch grid and pixel statistics:

\begin{itemize}
\tightlist
\item
  a dark, low-contrast patch near the left border of the key image, 15 patches from the caliper mark;
\item
  a perfectly uniform patch in the top row of the second image, with 105 identical copies in that image;
\item
  a perfectly uniform patch in the top row of the third image, with 212 identical copies;
\item
  a low-contrast interior patch with no identical copy.
\end{itemize}

\textbf{Long text, per layer} (Gemma-4-12B, five text-only prompts padded to about 2,260 tokens). The bf16--fp32 cosine of
the gradient at the output of layer 21 is 0.59--0.98 and falls gradually toward the lower layers; only one prompt
shows the abrupt drop at layers 19--20 that image prompts show, and two are already at 0.59 and 0.71 at layer 21. The
disagreement on long text therefore starts in part above layer 21.

\textbf{Long text on Gemma-4-31B} (lowest block, five prompts, 4-bit weights): 0.19--0.96 for short prompts (about 210
tokens), −0.41 to +0.40 with 2,170 tokens of diverse padding, and −0.13 to +0.07 when the padding is one sentence
repeated 140 times.

\section{Forward ablation}\label{forward-ablation}

To see what the detached path contributes to the forward pass, we set MLP outputs to zero in the base model (no
fine-tuning, bf16) and measured the NLL of the 80 validation answers of §5 (Table 8). At image tokens, the MLPs of
layers 19--21 and of the other mid-depth layers can be removed at little cost, and removing some of them lowers the
NLL slightly. The MLPs of early and late layers at image tokens, and of layers 19--21 at text tokens, cannot. This is
one task and an untuned model; it does not show that these MLPs are unused in general.

\begin{longtable}[]{@{}
  >{\raggedright\arraybackslash}p{(\linewidth - 8\tabcolsep) * \real{0.2000}}
  >{\raggedright\arraybackslash}p{(\linewidth - 8\tabcolsep) * \real{0.2000}}
  >{\raggedright\arraybackslash}p{(\linewidth - 8\tabcolsep) * \real{0.2000}}
  >{\raggedright\arraybackslash}p{(\linewidth - 8\tabcolsep) * \real{0.2000}}
  >{\raggedright\arraybackslash}p{(\linewidth - 8\tabcolsep) * \real{0.2000}}@{}}
\caption{\textbf{Validation NLL of the base model with MLP outputs zeroed in the forward pass.}}\tabularnewline
\toprule\noalign{}
\begin{minipage}[b]{\linewidth}\raggedright
MLP output zeroed at
\end{minipage} & \begin{minipage}[b]{\linewidth}\raggedright
In
\end{minipage} & \begin{minipage}[b]{\linewidth}\raggedright
Val. NLL
\end{minipage} & \begin{minipage}[b]{\linewidth}\raggedright
Change
\end{minipage} & \begin{minipage}[b]{\linewidth}\raggedright
Examples that get worse
\end{minipage} \\
\midrule\noalign{}
\endfirsthead
\toprule\noalign{}
\begin{minipage}[b]{\linewidth}\raggedright
MLP output zeroed at
\end{minipage} & \begin{minipage}[b]{\linewidth}\raggedright
In
\end{minipage} & \begin{minipage}[b]{\linewidth}\raggedright
Val. NLL
\end{minipage} & \begin{minipage}[b]{\linewidth}\raggedright
Change
\end{minipage} & \begin{minipage}[b]{\linewidth}\raggedright
Examples that get worse
\end{minipage} \\
\midrule\noalign{}
\endhead
\bottomrule\noalign{}
\endlastfoot
Nothing (base model) & & 0.653 & & \\
image tokens & layers 19--21 & 0.697 & +0.044 & 49 of 80 \\
text tokens & layers 19--21 & 1.331 & +0.678 & 80 of 80 \\
all tokens & layers 19--21 & 1.349 & +0.696 & 80 of 80 \\
image tokens & layers 4--6 & 1.274 & +0.621 & 80 of 80 \\
image tokens & layers 10--12 & 0.735 & +0.082 & 58 of 80 \\
image tokens & layers 13--15 & 0.663 & +0.011 & 38 of 80 \\
image tokens & layers 16--18 & 0.679 & +0.026 & 41 of 80 \\
image tokens & layers 22--24 & 0.692 & +0.040 & 43 of 80 \\
image tokens & layers 28--30 & 0.608 & −0.045 & 23 of 80 \\
image tokens & layers 40--42 & 1.030 & +0.377 & 80 of 80 \\
image tokens & layer 19 & 0.656 & +0.003 & 43 of 80 \\
image tokens & layer 20 & 0.595 & −0.058 & 22 of 80 \\
image tokens & layer 21 & 0.651 & −0.002 & 38 of 80 \\
\end{longtable}

\end{document}